# Compressing Deep Neural Networks Using Explainable AI

Kimia Soroush
*School of Electrical and Computer Engineering*
*Shiraz University*
Shiraz, Iran
soroush@shirazu.ac.ir

Mohsen Raji
*School of Electrical and Computer Engineering*
*Shiraz University*
Shiraz, Iran
mraji@shirazu.ac.ir

Behnam Ghavami
*Department of Computer Engineering*
*Shahid Bahonar University of Kerman*
Kerman, Iran
ghavami@uk.ac.ir

*Abstract*— **Deep neural networks (DNNs) have demonstrated remarkable performance in many tasks but it often comes at a high computational cost and memory usage. Compression techniques, such as pruning and quantization, are applied to reduce the memory footprint of DNNs and make it possible to accommodate them on resource-constrained edge devices. Recently, explainable artificial intelligence (XAI) methods have been introduced with the purpose of understanding and explaining AI methods. XAI can be utilized to get to know the inner functioning of DNNs, such as the importance of different neurons and features in the overall performance of DNNs. In this paper, a novel DNN compression approach using XAI is proposed to efficiently reduce the DNN model size with negligible accuracy loss. In the proposed approach, the importance score of DNN parameters (i.e. weights) are computed using a gradient-based XAI technique called Layer-wise Relevance Propagation (LRP). Then, the scores are used to compress the DNN as follows: 1) the parameters with the negative or zero importance scores are pruned and removed from the model, 2) mixed-precision quantization is applied to quantize the weights with higher/lower score with higher/lower number of bits. The experimental results show that, the proposed compression approach reduces the model size by 64% while the accuracy is improved by 42% compared to the state-of-the-art XAI-based compression method.**

## I. INTRODUCTION

In recent years, deep neural networks (DNNs) have been widely utilized in various electronic devices and services, from smartphones and home appliances to drones, robots, and self-driving cars. However, the problem arises when DNN models become significantly over-parameterized to increase their accuracy. So, the model size makes it impossible to deploy DNNs on the resource-constrained devices with limited memory and power consumption. Therefore, it requires rethinking the design, training, and deployment of DNNs in order to provide optimal balance between accuracy and other parameters such as memory and power consumption.

DNN compression techniques such as pruning and quantization have shown promising outcomes in accommodating DNNs with less memory footprint while providing acceptable accuracy [11]. In pruning, the goal is to remove some of the less important parameters, so it will not decrease the accuracy because the main weights or neurons in decision-making are preserved in the DNN model [12]. Quantization involves replacing datatypes with reduced width datatypes; for example, replacing 32-bit Floating Point (FP32) with 8-bit Integers (INT8). It allows speeding up inference and performing computations and storing tensors at lower than floating-point precision which is particularly beneficial during model deployment because the values can often be encoded to preserve more information than simple conversion [13]. In compression techniques, the main point is to find which components are unnecessary or could be preserved with less precision in order to decrease the model size while keeping the accuracy high.

There are several previous works on compressing the DNN models. In [8], Taylor decomposition is utilized to approximate the change in the cost function induced by pruning network parameters. However, it requires calculating the inverse of Hessian matrix that is computationally expensive. A method based on a per-layer pruning process is proposed in [10]. Quantization methods are introduced in [13] that reduce the energy and latency requirements of neural networks during inference. In [9], the importance of convolutional kernels is evaluated by the mean gradient criterion. A combination of pruning and quantization is proposed in [14], to achieve further network compression. These methods mostly rely in the parameter values to find the candidate to be pruned and/or quantized with lower precision. In [21], a compression framework based on merging weight pruning and quantization has been introduced which provides an efficient solution in terms of both time and space utilization. This innovation enhances practical performance compared to conventional post-training methods. However, the low-value weights are not necessarily the low-important ones in the final model output [1].

There are other metrics to use as a pruning or quantization. In [22], the Neuron Importance Score Propagation (NISP) algorithm is introduced which focuses on propagating importance scores of final responses just before the softmax classification layer in the network. This method involves a per-layer pruning process, which differs from our proposed metric since it does not take into account the overall global importance within the network. Additionally, Luo et al. [23] present ThiNet, a statistical channel pruning technique driven by data and based on statistics computed from the next layer. There are also hybrid approaches like the one in [24], which suggests a fusion method to combine weight-based channel pruning and network quantization. Also, Dai et al. [25] introduce an evolutionary paradigm that combines weight-based pruning with gradient-based growing to heuristically reduce network complexity.

The basic core of quantization and pruning-based compression techniques is the use of an *"important cereticia"* to determine which components are unneeded or might be kept with less precision in order to reduce model size while maintaining accuracy. However, traditional compression methods [15][16][17][18][19] which do not consider the "global importance" of the parameters in the model performance may misguide the compression algorithm in finding less important weights for pruning or quantizing, leading to inefficient compressed DNN models.

Explainable artificial intelligence (XAI) is a new research area devoted to understand and explain AI methods. XAI can be utilized to get to know the inner functioning of DNNs, such as the importance of different neurons and features in the overall performance of DNNs. Sabih et al. [3] use an XAI method called DeepLIFT [6] to calculate the importance score of weights to compress unimportant weights. However, the main challenge with this XAI method is that it needs to choose a reference activation that is important for the outcome of the compression method and often requires domain-specific knowledge. Gradient-based XAI methods are another category of XAI techniques in which no reference activation is required to find the importance of the score. A gradient-based XAI method called Layer-wise Relevance Propagation (LRP) is used in [2] to prune unimportant weights. However, this work considers only pruning for compressing DNNs. In [1], ECQˣ is proposed in which LRP is used to cover the sparsification due to the entropy-based quantization and reassign weights that were mistakenly assigned to zero. Explainability-centric compression methods employed in the past have encountered significant shortcomings. They often fall into one of two categories: those reliant on XAI techniques, which demand supplementary data to determine importance scores, or those limited to pruning or applying single-precision quantization for compressing deep neural networks (DNNs).

In this paper, a novel DNN compression approach is proposed based on LRP which a gradient-based XAI method. In the proposed XAI-based compression approach, the importance scores obtained through LRP is used as a metric for pruning and mixed precision quantization. The original DNN model in which the weights are represented in full precision format is given as the input the compression algorithm. The XAI method (i.e. LRP) is used to calculate the effect of the input on the resulting output as the importance scores. The obtained scores from the sample color image is demonstrated as another image in red, black, and white; i.e. red spots are where the importance score is positive while grey spots are where the score is negative meaning they do not affect the network output. Then, based on the obtained scores, unimportant neurons are removed, the medium-value-score neurons are stored with less precision and the most ones are quantized with higher number of bits. Experimental results show that, the proposed LRP-based pruning and mixed precision quantization decrease the model size by 64% and improves the accuracy about 42% compared to the state-of-the-art XAI-based compression method.

The rest of this paper is organized as follows: Section II presents the proposed compression method which is based on a gradient-based XAI method. Section III shows the experimental results and finally, section IV concludes the paper.

## II. Gradient-based XAI for Pruning and Mixed-precision Quantization of DNNs

In this section, the proposed compression method is explained. The final output of the algorithm is the importance scores that can recognize specifically the places where pruning and quantization should be done.

### A. The proposed compression criterion

Choosing a suitable compression criterion is an important step in doing pruning and quantization. LRP is a gradient-based XAI method based on the conservation of flows and proportional decomposition [5]; i.e. it means that if there is explainable evidence in the output, it must show up somewhere in the input features. Among all XAI algorithms, LRP has a clearly defined meaning, namely the contribution of an individual network unit, weight or filter, to the network output.

In the proposed compression technique, LRP is applied to extract the importance (relevance) scores of individual neurons and filters. LRP is aligned to the layered structure of machine learning models. In a model with $n$ layers, LRP calculates the activations of all layers, and then, during the backward pass, the network output is redistributed to all units of the network in a layer-by-layer fashion. LRP is based on a conservation property, where what has been received by a neuron must be redistributed to the lower layer in equal amounts.

In order to understand the process of calculating the importance scores of neurons, considering a layer's output neuron j, the distribution of its assigned importance score $R_j$ towards its lower layer input neurons i can be achieved by this rule:

$$R_{i \leftarrow j} = \frac{z_{ij}}{z_j} R_j \qquad \text{Eq. (1)}$$

where $z_j$ is the sum of pre-activations $z_{ij}$ at output neuron j; i.e.:

$$z_j = \sum_j z_{ij} \qquad \text{Eq. (2)}$$

where $z_{ij}$ is the contribution of neuron i to the activation of neuron j. In order to enforce the conservation principle over all i contributing to j, we have:

$$\sum_i R_{i \leftarrow j} = R_j \qquad \text{Eq. (3)}$$

where the importance score is shown using $R_{i \leftarrow j}$, and it shows how much the effect of output neuron j assigned importance score $R_j$ towards its lower layer input neurons i can be achieved by applying the above decomposition rule.

Neuron scores are calculated using the following formula:

$$R_{i \leftarrow j} = a_i w_{ij} \frac{R_j}{z_j}$$
$$where \quad a_i w_{ij} = a_i \frac{\partial z_j}{\partial a_i} \qquad \text{Eq.(4)}$$
$$is \ replaced \ by \quad w_{ij} \frac{\partial z_j}{\partial w_{ij}}$$

After having the score value of neuron i, (i.e. $R_i$), a decision must be taken about which filters should be kept. Based on the sorted array of score values, unnecessary filters are removed from the network because of low-score.

### B. The Compression Algorithm

Algorithm 1 presents the proposed LRP-based compression method. The inputs of the proposed method are the original model and the bit-widths that are going to be used in mixed-precision quantization. The first step is to calculate filter scores (Algorithm 1, lines 1 to 3). The filters which have negative score values and are not considered essential units, so removing them, will not affect the final result (Algorithm 1 , line 5). After removing the negative or zero score filters, the remaining filters have positive values and their importance score indicates a stronger contribution to the overall neural network accuracy or performance. Therefore, pruning them could decrease the accuracy, but they do not have that much contribution that we keep all of them with full precision. In this step, filters should be divided into two groups based on their importance scores, so a threshold is needed to classify them. The threshold is the median of the sorted scores in this model, because it is easily understood and is easy to calculate and also is not at all affected by extreme values. In this regard, other metrics like weighted mean where tested but the best accuracy was related to the experiment which classification was done using median. As mentioned before, according to the layer-by-layer approach, median of each layer is different from the other. Therefore, median of each layer is calculated and the weights with lower-than-median filter scores are assigned to a lower-bit-width value, and the others (as of their importance) are assigned to a higher bit-width (Algorithm 1 lines 7 to 13). The advantage of classifying filters based on their importance scores is that the unnecessary neurons are identified, so their weights are not kept. In addition, instead of keeping all the remaining weights, which are essential components, with full precision, we assign storage space directly related to their importance score. In this way, there is no extra storage used by the network.

## III. Experimental Results

In this section, the proposed compression method is evaluated using a DNN model and a dataset which is applicable in embedded systems that have a low-storage memory and limited computational power. In the following, the experimental setup is presented first and then, the comparison between state-of-art pruning methods and also non-XAI-based ones are explained in the following sub-sections, respectively.

---

**Algorithm 1** Explainability-Driven Quantization and Pruning

**Input:** Unpruned model, highBitWidth, lowBitWidth
**Output:** Quantized model
1: **for** all filters in a network **do**
2:      Calculate filters' scores based on their LRP scores
3: **end for**
4: Sort scores
5: Remove filters with negative score values (Pruning)
6: $\tau$ = score values' median
7: **for** the remaining filters **do**
8:      **if** $filter\ score > \tau$ **then**
9:          Quantize its weights to highBitWidth
10:      **end if**
11:      **else**
12:          Quantize its weights to lowBitWidth
13: **end for**

---

### A. Experiment

All the experiments are conducted using the PyTorch deep learning framework, version '1.12.1+cu113' with Torchvision '0.13.1+cu113'. Also, we use Intel(R) Xeon(R) CPU @ 2.20GHz with 2 GB memory as an inference device.

The proposed method is used to compress a DNN model which is used by the previous works to show the efficacy of their XAI-based compression technique [2]. The used model consists of a sequence of three consecutive ReLU-activated dense layers which has 1000 hidden neurons. Also, the proposed method is trained and tested on a k-class toy dataset "multi" (k = 4) using respective generator functions to investigate whether the suggested algorithm works well if only a minimal number of samples is available for compressing the model.

### B. LRP-based mixed-precision quantization

In order to show efficiency of LRP scores, the results of multiple experiments are shown in Table I. The first column indicates the method and each row indicates the performance of the method in terms of accuracy and model size. At first, some filters in each layer are removed so the remaining allocated storage is about 3.3 MB. In this stage, the accuracy is 91%, and 2000 filters out of 3000 filters remain. In the single-precision quantization (SP Q) done after pruning, all the weights in the remaining filters are quantized to a fixed bit-width regardless of their importance scores which are calculated by the LRP algorithm. Best accuracy in this situation is 90.8%. However, the better approach is to quantize weights to mixed-precision (MP Q) because less storage can be used in this situation. It means that in the remaining filters, the weights, which are related to the filters that have a lower-than-median score, are quantized to a lower bit-width because they have less effect on the result and could be saved with less accuracy while not that important that saving them with less precision could hurt the result. The weights in the remaining filters, which have the most effect on the output accuracy and highest importance scores, are quantized to a higher bit-width. Our other contribution is to quantize weights based on the layers. Different mixed-precision bit-widths are tried in order to find the best performance. We added this flexibility to our work so that there is no need to store the weights of all layers with the same accuracy.

TABLE I.     MODEL SIZE AND ACCURACY COMPARISON OF MODEL AFTER APPLYING THE COMPRESSION TECHNIQUE.
(SP Q: SINGLE PRECISION QUANTIZATION, MP Q: MIXED-PRECISION QUANTIZATION)

| Method | Pruning | Quantization | Model Size (Mb) | Accuracy |
|---|---|---|---|---|
| Original | - | - | 8 MB | 94.1% |
| Pruning | ✓ | - | 3.3 MB | 91% |
| Pruning + SP Q | ✓ | SP Q(16 bits) | 1.65 MB | 90.8% |
|  |  | SP Q(16 bits) | 0.82 MB | 90.1% |
| Pruning + MP Q | ✓ | Layer0: 8,16 bits | 0.9 MB | 89.8% |
|  |  | Layer1: 8,16 bits |  |  |
|  |  | Layer2: 4,8 bits |  |  |
| Pruning + MP Q | ✓ | Layer0: 8,16 bits | **1.2 MB** | **90.5%** |
|  |  | Layer1: 8,16 bits |  |  |
|  |  | Layer2: 8,16 bits |  |  |

Figure 1 shows the comparison of different XAI-based compression approaches in terms of accuracy after compression. In this figure, the vertical axis shows the method and the horizontal axis shows the final model accuracy. It is evident that our proposed method can relatively improve the accuracy by 42% in comparison to the algorithm when pruning and single-precision quantization (SP Q) is applied. The better accuracy validates the idea that Explainability-Driven approaches can recognize important neurons correctly. Note that the previous works used LRP to prune unimportant weights, while the algorithm proposed in this paper used the scores in all layers to quantize the weights.

Figure 2 compares of the proposed compression method with its counterparts in terms of model size. The vertical axis shows the method and the horizontal axis shows the model size. Based on this approach, the model size relatively decreases by 64%, which is considerable when deployed to embedded devices. This result shows how using LRP scores to quantize weights with different precision helps the compression while not hurting the accuracy.

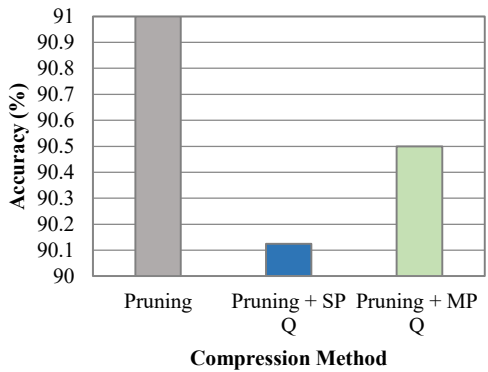

Fig. 1. Accuracy comparison of the proposed method (Pruning + MP Q) and other criteria (Pruning and Pruning + SP Q)

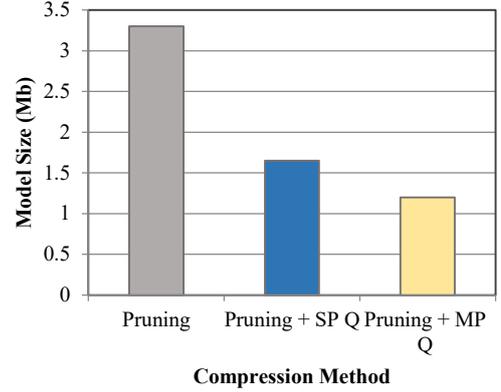

Fig. 2. Model Size comparison of the proposed method (LRP + MP Q) and other approaches (Pruning and Pruning + SP Q)

## C. Comparison of XAI-based vs non-XAI-based compression method

The superiority of the proposed XAI-based compression approach is shown by a comparison to a non-XAI-based compression method. Table II shows the performance of the proposed XAI-based pruning and mixed-precision quantization method compared to a non-XAI-based approach [8]. In this experiment, the non-XAI-based algorithm is a first-order Taylor expansion technique that estimates the change of loss in the objective function due to pruning away network units and quantizing weights. Putting aside its low accuracy when the quantization is applied, computing the inverse of Hessian is computationally expensive. According to the obtained results, the accuracy improved about 3% when using explainability which means its ability to find important model components is high enough to be used as a metric for compression. The single 8-bit precision quantization using Taylor decomposition is compared to mixed-precision (8 bits for less important weights and 16 bits for more important weights) quantization using LRP scores are done.

TABLE II.    PERFORMANCE COMPARISON BETWEEN XAI-BASED ALGORITHM VS. A NON-XAI-BASED ONE ALGORITHMS

| Method | Model Size | Accuracy | |
|---|---|---|---|
| | | Non-XAI | XAI |
| Pruning | 3.3 MB | 88% | 91% |
| Pruning + SP Q | 0.82 MB | 87% | 90.125% |
| Pruning + MP Q | 1.2 MB | **87%** | **90.5%** |

### D. Comparison with the stae-of-the art

In order to compare the proposed method with the state-of-the-art works, we choose the LRP-based filter pruning method proposed in [2] in which less significant filters are pruned from a DNN to reduce its size. We also compare the proposed method with the weight pruning method presented in [21]. The obtained results is reported in Table III where the first column shows the compression schemes (only pruning, pruning and single precision quantization (Pruning + SP Q), and pruning and multi-precision quantization (Pruning +MP Q)) while the second column indicate the model size achieved with the compression scheme. The accuracy obtained by different methods of evaluating importance scores are presented in the next columns. As shown in this table, the proposed method achieves more than 90% accuracy in case of Pruinng + MP Q while the accuracy of the other two methods is less than 80%. This superiority is originated from using LRP-based weight importance metric while the method in [21] only consider the weight magnitude as the metric of the important metric and the presented approach in [2] assigns filter relevance value for weight importance value and does not calculate relevances of each weight which leads to degradation in the accuracy of the compressed network.

TABLE III.    PERFORMANCE COMPARISON BETWEEN THE PROPOSED METHOD VS. STATE-OF-THE ART ALGORITHM

| Method | Model Size | Accuracy | | |
|---|---|---|---|---|
| | | Weight Magnitude[21] | Filter pruning[2] | Proposed Method |
| Pruning | 3.3 MB | 80.30% | 88.25% | 91% |
| Pruning+ SP Q | 0.82 MB | 75.45% | 82.01% | 90.125% |
| Pruning+ MP Q | 1.2 MB | 71.49% | 79.25% | 90.5% |

### IV.    CONCLUSION

This paper introduced a novel mixed-precision method that utilizes weight relevance information from Layer-wise Relevance Propagation (LRP) to compress deep neural networks. Unnecessary filters in the decision-making of the model have been pruned. Among the remaining neurons, more important ones are preserved with a bit-width near full precision, while others are stored with less accuracy. LRP-based classification has shown considerable compression and high accuracy compared to full-precision mode. The model size is reduced to 36% of its original model and its accuracy improved to 42% compared to previous work. Our results have shown the outstanding performance of Explainable-AI in understanding the decisions of neural networks and their application in compression. Therefore, we propose using this algorithm for future work to calculate other components' importance scores and minimize the network size to be applicable in embedded devices.